\pdfoutput=1

\documentclass[11pt]{article}

\usepackage[]{acl}

\usepackage{times}
\usepackage{latexsym}

\usepackage[T1]{fontenc}

\usepackage[utf8]{inputenc}

\usepackage{microtype}

\usepackage{graphicx}
\usepackage{amsmath}
\usepackage{amssymb}
\usepackage{booktabs}
\usepackage[linesnumbered,ruled,vlined]{algorithm2e}
\usepackage{multirow}

\title{What changes when you randomly choose BPE merge operations?\\ Not much.$^*$}

\author{Jonne S{\"a}lev{\"a} \and Constantine Lignos \\
  Michtom School of Computer Science \\
  Brandeis University \\
  \texttt{\{jonnesaleva,lignos\}@brandeis.edu}}

\begin{document}

\maketitle
\begin{abstract}
We introduce three simple randomized variants of byte pair encoding (BPE) and explore whether randomizing the selection of merge operations substantially affects a downstream machine translation task. We focus on translation into morphologically rich languages, hypothesizing that this task may show sensitivity to the method of choosing subwords. Analysis using a Bayesian linear model indicates that two of the variants perform nearly indistinguishably compared to standard BPE while the other degrades performance less than we anticipated. We conclude that although standard BPE is widely used, there exists an interesting universe of potential variations on it worth investigating. Our code is available at: \url{https://github.com/bltlab/random-bpe}.
\end{abstract}



\def\thefootnote{*}\footnotetext{
This version of the paper contains an additional experimental condition (the count-proportional BPE variant) that does not appear in the version in the ACL Anthology.
}\def\thefootnote{\arabic{footnote}}

\section{Introduction and related work}

Most neural machine translation (NMT) models assume their inputs to be sequences of units drawn from a fixed vocabulary.
While these units were tokens in the early years of NMT \citep{cho-etal-2014-learning,sutskever2014sequence}, 
there has since been a transition to \textit{subword}-level models that learn a vocabulary of ``word pieces'' which serve as an intermediate representation between words and characters \citep{mielke2021between}.
Such representations are attractive because they solve the closed-vocabulary problem of early, word-level NMT \citep{luong-etal-2015-effective} while also yielding more semantically meaningful units than individual characters.

Well-known subword segmentation algorithms include byte pair encoding (BPE) \citep{sennrich-etal-2016-neural},  SentencePiece Unigram LM \citep{kudo-richardson-2018-sentencepiece, kudo-2018-subword} and the WordPiece algorithm \citep{wu2016wordpiece,song-etal-2021-fast}. 
All of them include a hyperparameter that controls the size of the subword vocabulary: SentencePiece and WordPiece do this explicitly with a vocabulary size parameter, whereas BPE specifies the number of \textit{merge operations} which implicitly define the subword vocabulary.

Prior work has addressed the problem of optimally selecting the vocabulary size. 
\citet{haddow-etal-2018-university} and \citet{sennrich-zhang-2019-revisiting} find that using too large a subword vocabulary can result in low-frequency tokens being represented as atomic units, which makes it difficult to learn proper representations for them. \citet{gowda-may-2020-finding} suggest a heuristic: use as many subwords as possible provided that at least 95\% of the subwords have 100 or more examples in the training set.
\citet{gutierrez-vasques-etal-2021-characters} find that around 350 merge operations are enough to generate similar subword distributions across languages.

Subword segmentation algorithms usually build their subword vocabularies by optimizing an objective function that is independent of the downstream task. For instance, SentencePiece employs the probabilities under its unigram language model, while BPE aims to maximize the degree of sequence compression by greedily selecting and merging the symbol pairs that occur most frequently.
Others have re-framed this process as finding an ``optimal'' set of units that maximize more sophisticated probabilistic criteria. 
\citet{vilar-federico-2021-statistical} introduce an extension of BPE that learns a subword vocabulary by maximizing a likelihood objective over potential subwords. \citet{he-etal-2020-dynamic} introduce a method that treats the segmentation as a latent variable to be marginalized out and seek to find segmentations that maximize the downstream task probability directly.

In this paper, we build upon the concept of stochastic segmentation and conduct neural machine translation experiments on four languages (German, Finnish, Estonian and Uzbek) of varying morphological complexity, using variants of BPE that randomly sample merge operations instead of deterministically choosing the most frequent one.

Our negative result challenges our initial beliefs that standard BPE would produce the most effective subword representations for translation and that the success of BPE was due to the greedy selection process for learning merge operations.
We find that even when merge operations are randomly sampled uniformly, the performance degradation is less than we anticipated.
We conclude by discussing how this finding relates to the overall role of subwords in NMT.

\section{Byte pair encoding and randomization}

We briefly review the BPE training algorithm and introduce our randomized variants.
The pseudocode for the algorithm we use can be seen in Algorithm~\ref{bpe_train_algo}.
Our presentation is adapted from the BPE algorithm in \citet{vilar-federico-2021-statistical}.

\begin{algorithm}[tbh]
\footnotesize
\DontPrintSemicolon
\SetKwInput{KwInput}{Input}
\SetKwInput{KwOutput}{Output}
\SetKwComment{Comment}{/* }{ */}

\KwInput{$D$: Training corpus. $M$: Number of merge operations to learn.}
\KwOutput{$R$: list of learned merges.}

\SetKwFunction{FTrain}{trainBPE}
\SetKwProg{Fn}{def}{:}{}

\Fn{\FTrain{$D$, $M$, method}}{
    $R\gets []$\;
    \While{$|R| \leq M$}{
        $\text{C} \gets \text{countSymbolPairs}(D)$\;
        $(x,y) \gets \text{choosePair}(\text{C}, \text{method})$\;
        $\text{rule} \gets \langle(x,y) \rightarrow xy\rangle$\;
        $R \gets \text{append}(R, \text{rule})$\;
        $D \gets \text{applyRule}(D, \text{rule})$\;

    }
    \Return $R$
}

\SetKwFunction{FChoose}{choosePair}
\SetKwProg{Fn}{def}{:}{}
    \Fn{\FChoose{$\text{counts, method}$}}{
    \uIf{method = standard}
    {
        $\text{pair} \gets \arg\max_{\text{pair} \in \text{counts}} \text{counts}[\text{pair}]$
    }
    \uElseIf{method = softmax}{
        $\text{probs} \gets \text{softmax}(\text{counts})$ \;
        $\text{pair} \gets \text{sample}(\text{counts}, \text{probs})$ \;
    }
    \uElseIf{method = countprop}{
        $\text{probs} \gets \text{counts}/\text{sum}(\text{counts})$ \;
        $\text{pair} \gets \text{sample}(\text{counts}, \text{probs})$ \;
    }
    \uElseIf{method = uniform}{
        $\text{probs} \propto 1$  \;
        $\text{pair} \gets \text{sample}(\text{counts}, \text{probs})$ \;
    }    
    \Return $\text{pair}$
}
    \caption{BPE training algorithm.}

\label{bpe_train_algo}
\end{algorithm}

\subsection{Standard BPE algorithm}

The standard byte pair encoding algorithm \citep{sennrich-etal-2016-neural} is a greedy algorithm that takes as input a corpus $D$---typically the training set or another large collection of text---as well an integer $M$ that specifies the number of merge operations to learn.
After first segmenting $D$ into space-separated characters, the algorithm counts how many times each pair of symbols occurs in $D$ (\texttt{countSymbolPairs}). Based on the counts, the algorithm finds the most frequent symbol pair (\texttt{choosePair}) and learns a new merge operation that merges the constituent symbols into a new symbol.
After learning the merge operation, the algorithm replaces all occurrences of the symbol pair in $D$ with the new merged symbol (\texttt{applyRule}).

While the initial merge operations merge individual characters, during the later iterations larger chunks of words are merged together as well.
For example, if the most frequent symbol pair was \texttt{(ab,c)}, the algorithm would learn the rule \texttt{ab~c} $\rightarrow$ \texttt{abc} which replaces all occurrences of \texttt{ab~c} with \texttt{abc}, taking care to not cross word boundaries.

This is repeated for $M$ iterations until a desired number of merge operations is learned, after which the algorithm returns the list of merge operations as output.
At test time, the algorithm splits incoming lines of text into individual characters and then applies each of the learned merge operations in order, resulting in text where each space-separated token is an individual subword.

\subsection{Randomized BPE variants}
To extend BPE to randomized variants, we replace the step of picking the most frequent symbol pair at each iteration with random sampling.

\paragraph{Softmax sampling}
In our first variant, we assign each symbol pair a probability of being sampled based on how often it occurs in the observed data.
We apply a softmax to the observed symbol pair occurrence counts and draw a random symbol pair to merge according to a categorical distribution with the softmax probabilities as its parameters.

\paragraph{Count-proportional sampling}
In our second variant, we sample symbol pairs with probability equal to the normalized count of each pair.
Instead of using a softmax to obtain probabilities, we simply divide each count by the sum of all counts which yields a less peaked distribution than the one induced by a softmax.
Random sampling proceeds in the same way as with softmax sampling using a categorical distribution parameterized by the normalized counts.

\paragraph{Uniform sampling}
As our last variant, we select each merge operation with uniform probability from the set of observed symbol pairs.
Since every symbol pair has equal probability of being sampled, the frequency of each symbol pair is not used in sampling.

\section{Experimental setup}

\begin{table*}[tb]
\centering
\resizebox{\linewidth}{!}{
\footnotesize
\begin{tabular}{llcccccccc}
\toprule
      &  & \multicolumn{4}{c}{BLEU} & \multicolumn{4}{c}{CHRF} \\
           \cmidrule(lr){3-6} \cmidrule(lr){7-10}
Language & Merges  & Standard & Softmax & CountProp & Uniform & Standard & Softmax & CountProp & Uniform \\ 
\midrule
\multirow{3}{*}{Estonian} & 2,000  &  18.08 \tiny{(0.07)} &  18.17 \tiny{(0.06)} &  17.87 \tiny{(0.05)} &  17.48 \tiny{(0.04)} &  51.01 \tiny{(0.09)} &  51.10 \tiny{(0.09)} &  50.85 \tiny{(0.04)} &  50.38 \tiny{(0.07)} \\
& 5,000  &  17.98 \tiny{(0.11)} &  17.89 \tiny{(0.09)} &  17.52 \tiny{(0.07)} &  17.43 \tiny{(0.06)} &  50.80 \tiny{(0.14)} &  50.65 \tiny{(0.11)} &  50.45 \tiny{(0.07)} &  50.48 \tiny{(0.06)} \\
& 32,000 &  16.13 \tiny{(0.06)} &  16.13 \tiny{(0.10)} &  15.79 \tiny{(0.07)} &  16.86 \tiny{(0.06)} &  48.70 \tiny{(0.09)} &  48.65 \tiny{(0.05)} &  48.29 \tiny{(0.09)} &  50.21 \tiny{(0.09)} \\
\midrule
\multirow{3}{*}{Finnish} & 2,000  &  16.40 \tiny{(0.08)} &  16.20 \tiny{(0.04)} &  15.71 \tiny{(0.11)} &  15.26 \tiny{(0.14)} &  50.99 \tiny{(0.07)} &  50.90 \tiny{(0.06)} &  50.44 \tiny{(0.05)} &  49.76 \tiny{(0.12)} \\
& 5,000  &  15.77 \tiny{(0.08)} &  16.01 \tiny{(0.04)} &  15.26 \tiny{(0.06)} &  14.63 \tiny{(0.09)} &  50.64 \tiny{(0.07)} &  50.67 \tiny{(0.06)} &  50.04 \tiny{(0.10)} &  49.32 \tiny{(0.09)} \\
& 32,000 &  13.83 \tiny{(0.09)} &  13.88 \tiny{(0.09)} &  13.36 \tiny{(0.17)} &  13.28 \tiny{(0.11)} &  48.20 \tiny{(0.11)} &  48.20 \tiny{(0.07)} &  47.37 \tiny{(0.10)} &  47.92 \tiny{(0.08)} \\
\midrule
\multirow{3}{*}{German} & 2,000  &  24.56 \tiny{(0.05)} &  24.46 \tiny{(0.06)} &  23.95 \tiny{(0.07)} &  22.54 \tiny{(0.08)} &  55.77 \tiny{(0.03)} &  55.74 \tiny{(0.03)} &  55.42 \tiny{(0.02)} &  53.41 \tiny{(0.08)} \\
& 5,000  &  24.84 \tiny{(0.07)} &  24.79 \tiny{(0.10)} &  24.31 \tiny{(0.07)} &  22.73 \tiny{(0.04)} &  56.12 \tiny{(0.04)} &  55.98 \tiny{(0.04)} &  55.71 \tiny{(0.05)} &  53.65 \tiny{(0.05)} \\
& 32,000 &  25.49 \tiny{(0.06)} &  25.33 \tiny{(0.05)} &  24.94 \tiny{(0.05)} &  22.91 \tiny{(0.07)} &  56.60 \tiny{(0.03)} &  56.54 \tiny{(0.04)} &  56.23 \tiny{(0.05)} &  54.26 \tiny{(0.05)} \\
\midrule
\multirow{3}{*}{Uzbek} & 2,000  &  47.31 \tiny{(0.21)} &  45.82 \tiny{(1.14)} &  45.14 \tiny{(0.18)} &  37.85 \tiny{(0.24)} &  64.51 \tiny{(0.18)} &  63.24 \tiny{(1.00)} &  62.91 \tiny{(0.15)} &  57.66 \tiny{(0.23)} \\
& 5,000  &  46.77 \tiny{(1.10)} &  45.39 \tiny{(1.46)} &  47.40 \tiny{(0.19)} &  38.79 \tiny{(0.20)} &  63.78 \tiny{(0.92)} &  62.52 \tiny{(1.31)} &  64.45 \tiny{(0.17)} &  58.39 \tiny{(0.15)} \\
& 32,000 &  48.63 \tiny{(0.75)} &  47.98 \tiny{(0.70)} &  46.87 \tiny{(0.52)} &  41.73 \tiny{(0.51)} &  64.76 \tiny{(0.56)} &  64.24 \tiny{(0.59)} &  63.42 \tiny{(0.41)} &  60.43 \tiny{(0.42)} \\
\bottomrule
\end{tabular}
}
\caption{Mean and standard error of BLEU and chrF scores across target languages, merge operations and BPE segmentation types. All numbers computed over 10 replications with different random seeds.}
\label{tab:main_results}
\end{table*}

\paragraph{Task and data}

We experiment with translation from English to several morphologically rich languages: Finnish, Estonian, German, and Uzbek.
Statistics for each dataset can be found in the Appendix.
For all languages except Uzbek, we use the WMT shared task data from \citet{he-etal-2020-dynamic}. For Uzbek, we use the Turkic Interlingua corpus \citep{mirzakhalov-etal-2021-evaluating}.

\paragraph{Tokenization and subword segmentation}
All of our datasets had previously been tokenized.
We performed BPE segmentation on those tokens at the character level using \texttt{subword-nmt}, which we modified to support randomized subword sampling.
All subword vocabularies are learned separately for each language.
As the number of merge operations $M$ is a hyperparameter, we experiment with the values 2,000, 5,000, and 32,000.
The largest value, 32,000, is taken directly from \citet{he-etal-2020-dynamic};
the smaller values of 2,000 and 5,000 are motivated the observation that higher numbers of merges tend to lead to a near-word-level segmentations for which learning good representations may not be feasible \citep{sennrich-zhang-2019-revisiting}.

\paragraph{Model and training}
Our model is a standard Transformer-based encoder-decoder model, as implemented in the \texttt{fairseq} library.
Our architecture is similar to \texttt{transformer-base}, with 512-dimensional embeddings on both the encoder and decoder side, 2048-dimensional feedforward layers, and 6 stacked Transformer layers with 8 attention heads each in both the encoder and decoder.
We train all our models for 10,000 updates using a learning rate of 0.005 and the largest feasible batch size (36K tokens per batch for Finnish and Estonian, 30K tokens per batch for German, and 12K tokens per batch for Uzbek).
Each translation experiment is run on a single NVIDIA V100 GPU (24GB).
We simulate training on multiple GPUs by accumulating gradients for 16 backward passes before each parameter update.
To estimate the variability of our results across random seeds, we perform 10 replications of each experiment. 

\paragraph{Evaluation}
We evaluate all of our models with the \texttt{sacrebleu} library \citep{post-2018-call} using BLEU\footnote{Version string: \texttt{nrefs:1|case:mixed|eff:no|\\tok:13a|smooth:exp|version:2.3.1}} \citep{papineni-etal-2002-bleu} as well as chrF\footnote{Version string: \texttt{nrefs:1|case:mixed|eff:yes|\\nc:6|nw:0|space:no|version:2.3.1}} \citep{popovic-2015-chrf} as it is a tokenization-free metric.
Both metrics are computed using the default parameters.
We use the \texttt{sacremoses}\footnote{\url{https://github.com/alvations/sacremoses}} detokenizer to create the detokenized versions of our corpora.

\section{Results}

\begin{figure*}[tb]
    \centering
    \includegraphics[width=\textwidth]{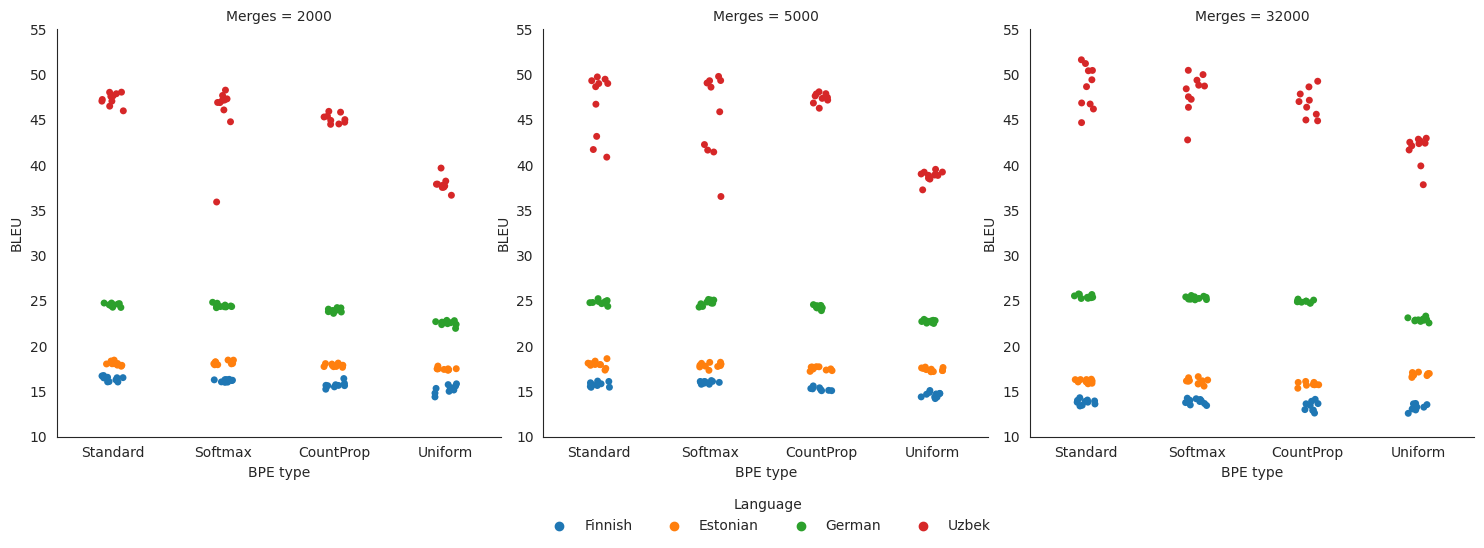}
    \caption{Translation performance (BLEU) across languages and merge operations. A figure showing chrF is provided in the Appendix.}
    \label{fig:bleu_results}
\end{figure*}

\begin{table}[tb]
    \centering
\resizebox{\linewidth}{!}{
\begin{tabular}{lrrrc}
\toprule
                           Parameter &  Mean &  HDI (lower) &  HDI (upper) & Spans zero? \\
\midrule
\multicolumn{3}{l}{\textit{BPE effects (average)}} & & \\
\midrule
     $\bar{\beta}_{\text{Standard}}$ &  0.56 &        -0.60 &         1.67 &  \checkmark \\
      $\bar{\beta}_{\text{Softmax}}$ &  0.26 &        -0.84 &         1.37 &  \checkmark \\
    $\bar{\beta}_{\text{CountProp}}$ & -0.04 &        -1.19 &         1.06 &  \checkmark \\
      $\bar{\beta}_{\text{Uniform}}$ & -0.98 &        -2.61 &         0.47 &  \checkmark \\
\midrule
\multicolumn{3}{l}{\textit{BPE effects (language-specific)}} & & \\
\midrule
 $\beta_{\text{Standard, Estonian}}$ &  0.41 &        -0.75 &         1.65 &  \checkmark \\
  $\beta_{\text{Standard, Finnish}}$ &  0.48 &        -0.73 &         1.64 &  \checkmark \\
   $\beta_{\text{Standard, German}}$ &  0.52 &        -0.64 &         1.74 &  \checkmark \\
    $\beta_{\text{Standard, Uzbek}}$ &  0.98 &        -0.25 &         2.21 &  \checkmark \\
  $\beta_{\text{Softmax, Estonian}}$ &  0.33 &        -0.86 &         1.49 &  \checkmark \\
   $\beta_{\text{Softmax, Finnish}}$ &  0.41 &        -0.84 &         1.55 &  \checkmark \\
    $\beta_{\text{Softmax, German}}$ &  0.36 &        -0.83 &         1.53 &  \checkmark \\
     $\beta_{\text{Softmax, Uzbek}}$ &  0.07 &        -1.08 &         1.26 &  \checkmark \\
$\beta_{\text{CountProp, Estonian}}$ & -0.00 &        -1.19 &         1.13 &  \checkmark \\
 $\beta_{\text{CountProp, Finnish}}$ & -0.08 &        -1.23 &         1.12 &  \checkmark \\
  $\beta_{\text{CountProp, German}}$ & -0.05 &        -1.24 &         1.10 &  \checkmark \\
   $\beta_{\text{CountProp, Uzbek}}$ & -0.01 &        -1.12 &         1.20 &  \checkmark \\
  $\beta_{\text{Uniform, Estonian}}$ &  0.20 &        -1.00 &         1.45 &  \checkmark \\
   $\beta_{\text{Uniform, Finnish}}$ & -0.50 &        -1.68 &         0.78 &  \checkmark \\
    $\beta_{\text{Uniform, German}}$ & -1.72 &        -2.93 &        -0.50 &             \\
     $\beta_{\text{Uniform, Uzbek}}$ & -6.92 &        -8.16 &        -5.69 &             \\
\midrule
\multicolumn{3}{l}{\textit{Merge effects (average)}} & & \\
\midrule
        $\bar{\gamma}_{\text{2000}}$ &  0.10 &        -1.13 &         1.36 &  \checkmark \\
        $\bar{\gamma}_{\text{5000}}$ &  0.15 &        -1.20 &         1.38 &  \checkmark \\
       $\bar{\gamma}_{\text{32000}}$ & -0.07 &        -1.38 &         1.32 &  \checkmark \\
\midrule
\multicolumn{3}{l}{\textit{Merge effects (language-specific)}} & & \\
\midrule
    $\gamma_{\text{2000, Estonian}}$ &  0.32 &        -1.05 &         1.72 &  \checkmark \\
     $\gamma_{\text{2000, Finnish}}$ &  0.48 &        -0.92 &         1.90 &  \checkmark \\
      $\gamma_{\text{2000, German}}$ & -0.05 &        -1.35 &         1.39 &  \checkmark \\
       $\gamma_{\text{2000, Uzbek}}$ & -0.25 &        -1.77 &         1.18 &  \checkmark \\
    $\gamma_{\text{5000, Estonian}}$ &  0.17 &        -1.24 &         1.52 &  \checkmark \\
     $\gamma_{\text{5000, Finnish}}$ &  0.10 &        -1.27 &         1.53 &  \checkmark \\
      $\gamma_{\text{5000, German}}$ &  0.18 &        -1.17 &         1.57 &  \checkmark \\
       $\gamma_{\text{5000, Uzbek}}$ &  0.17 &        -1.29 &         1.59 &  \checkmark \\
   $\gamma_{\text{32000, Estonian}}$ & -1.29 &        -2.67 &         0.18 &  \checkmark \\
    $\gamma_{\text{32000, Finnish}}$ & -1.72 &        -3.21 &        -0.27 &             \\
     $\gamma_{\text{32000, German}}$ &  0.69 &        -0.76 &         2.02 &  \checkmark \\
      $\gamma_{\text{32000, Uzbek}}$ &  1.93 &         0.36 &         3.31 &             \\
\bottomrule
\end{tabular}
}
\caption{Posterior means and 94\% posterior highest density intervals for the BLEU model.}
\label{tab:posterior_table}
\end{table}

Our main experimental results are displayed in Table~\ref{tab:main_results} and Figure~\ref{fig:bleu_results}.
For most languages translation performance appears to be rather stable across seeds, but in Uzbek standard errors are larger than other languages and they seem to increase with increasing numbers of merge operations.
We believe this noisiness is due to the smaller size of the Uzbek dataset rather than any language-specific phenomena.

Initially, we would have expected that standard BPE would perform the best out of all methods and that different BPE variants would produce noticeable performance differences for all languages.
Somewhat contrary to our hypothesis, we find that using randomized BPE variants seems to have quite a small average effect with significant variation in the effect size from language to language.
Uniform segmentation tends to consistently perform worse than standard BPE and softmax-based sampling, which can be explained by the roughly 3x longer sequences the model produces.
Looking across merge operations, the BLEU/chrF differences between the best and worst BPE variants seem to be less than 1.0--1.5 points for Estonian and Finnish, respectively.
However, German and Uzbek show a different picture, with BLEU/chrF differences of around 2-2.5 points for German and 4-9 points for Uzbek.

The impact of varying the number of merge operations varies and again bifurcates the set of languages. Finnish and Estonian seem to suffer slightly as the merge operations are increased (approx. -2 to -2.5 points in BLEU/chrF), whereas German and Uzbek seem to benefit from more merge operations (approx. +1 to +2 points in BLEU/chrF).
We find this perplexing, as the German and Uzbek datasets are the largest and smallest used in our experiments. 

To analyze these results, we fit a hierarchical, Bayesian linear model with language-specific effects for the BPE variant and number of merge operations:

\begin{equation*}
    \mu = \alpha^{(l)} + \beta_{b}^{(l)} + \gamma_{m}^{(l)} + \epsilon
\end{equation*} 

\noindent where $\alpha^{(l)}$ is an intercept, $\beta_{b}^{(l)}$ is the effect of using BPE variant $b$, $\gamma_{m}^{(l)}$ is the effect of using $m$ merge operations, and $\epsilon$ represents residual sampling error. All effects are specific to language $l$ and are drawn from common prior distributions: $\alpha^{(l)} \sim \mathcal{N}(0, \sigma^2_{\alpha})$, $\beta_{b}^{(l)} \sim \mathcal{N}(\bar{\beta}_b, \sigma^2_{\beta,b})$ and $\gamma_{m}^{(l)} \sim \mathcal{N}(\bar{\gamma}_m, \sigma^2_{\gamma,m})$.
Since our model is hierarchical, we also infer posteriors for the language-independent effects of using each BPE variant/number of merge operations, $\bar{\beta}_b$ and $\bar{\gamma}_m$, as well as the standard deviations $\sigma^2_{\beta,b}$ and $\sigma^2_{\gamma,m}$ that quantify between-language variation in the BPE and merge effects. We set the priors of the average effects to $\mathcal{N}(0, 1)$ and those of the standard deviations to $\mathcal{N}^{+}(1)$, except for $\sigma^2_{\alpha}$ for which we use the default $\mathcal{N}^{+}(s_{\alpha}), s_{\alpha} \approx 68$ prior specified by the Bambi modeling library \citep{bambi2022capretto} which we use to fit our model. We fit all our models using the No-U-Turn Sampler \citep{hoffmangelman2014nuts}. We run 4 Markov chains in parallel and draw 1,000 posterior samples from each chain. Prior to sampling, we also run each chain for 1,000 warm-up steps. 

Table~\ref{tab:posterior_table} shows a posterior mean point estimate for each effect of interest and quantifies their uncertainty using a 94\% highest density interval (HDI).
The effect sizes of randomized BPE variants seem confirm our experimental results. While the language-independent average effect sizes are all modest in magnitude, ranging from -0.98 for uniform BPE to +0.56 for standard BPE, there is substantial variation in the effect sizes when using uniform random sampling: effect sizes ranging from -6.92 for Uzbek to +0.20 for Estonian. 
Most importantly, the uncertainty intervals include zero for all languages and BPE types except for $\beta_{\text{Uniform, German}}$ and $\beta_{\text{Uniform, Uzbek}}$.

The effects for the number of merges are largely similar, with small average effects and between-language variation in effects on both sides of zero. 
While effect sizes tend to be very small for 2,000 and 5,000 merge operations, the effect varies with 32K merges. German and Uzbek seem to benefit from using 32K merge operations (posterior means +0.69 and +1.93, respectively). In contrast, Finnish and Estonian have significantly negative effect sizes (-1.72 and -1.29, respectively) with the entire 94\% HDI for Finnish below zero as well. 

\section{Discussion}

\subsection{Limitations and future work}

While our results suggest that randomized BPE segmentation algorithms have no consistent deleterious effect on BLEU/chrF across languages, it is possible that further experiments may find differently.
There is room for exploration regarding randomization of the BPE algorithm. 
For example, instead of sampling from the set of observed symbol pairs, merge operations could be chosen by sampling two unigrams independently or using a temperature-augmented sampler.

Although we focus on morphologically rich languages, our experiments still utilize a moderate amount of training data. 
Many morphologically rich languages that we did not consider may also lack such resources and thus be more impacted by the choice of subword segmentation algorithm. 
We feel that future work should pay particular attention should to this intersection of morphological complexity and low-resourcedness.


\subsection{Conclusion}

We introduced three randomized variants of BPE with the expectation that they would have a negative effect on translation performance because the traditional greedy approach should result in better subwords.
Instead, our results indicate that subword vocabularies created with randomized BPE yield translation models that perform comparably to those that use subwords created using the standard greedy BPE algorithm. Even when using uniform sampling, performance only degrades substantially for two of the languages we consider. This finding is corroborated by further analysis using a Bayesian linear model which suggests that the effect of uniform sampling is significantly different from zero for only German and Uzbek.

We find this negative result significant, as it suggests that variations on standard BPE can perform reasonably well.
We emphasize, however, that it is not clear whether this holds universally, particularly when using Transformer architectures optimized for handling longer sequences or when working with extremely small amounts of training data.
We hope that our negative result can motivate further research into the optimal use of subword segmentation algorithms, especially in the context of languages that are both morphologically rich and less-resourced, such as various Indigenous languages.

\bibliography{anthology,custom}
\bibliographystyle{acl_natbib}

\appendix

\section{Additional Tables and Figures}

\paragraph{Corpus statistics} 

\begin{table*}[ht!]
    \centering
    \footnotesize
    \resizebox{\textwidth}{!}{
\begin{tabular}{lllllllrr}
\toprule
      & {} &   & \multicolumn{2}{c}{Tokens} & \multicolumn{2}{c}{Types} & \multicolumn{2}{c}{Type-to-token ratio} \\
      \cmidrule(lr){4-5} \cmidrule(lr){6-7} \cmidrule(lr){8-9}
     Language & Split & Sentences    &     English & Non-English &  English & Non-English & English & Non-English \\
\midrule
\multirow{3}{*}{Estonian} & Train &  1,856,236 &  32,850,284 &  27,221,588 &  361,245 &     713,970 &    0.01 &        0.03 \\
      & Dev &      2,000 &      45,892 &      36,333 &    7,731 &      12,275 &    0.17 &        0.34 \\
      & Test &      2,000 &      48,340 &      38,063 &    8,085 &      12,956 &    0.17 &        0.34 \\
\midrule
\multirow{3}{*}{Finnish} & Train &  1,754,754 &  43,898,422 &  32,012,655 &  116,620 &     677,874 &    0.00 &        0.02 \\
      & Dev &      1,500 &      34,251 &      24,617 &    6,251 &      10,005 &    0.18 &        0.41 \\
      & Test &      1,370 &      29,183 &      21,142 &    5,761 &       8,958 &    0.20 &        0.42 \\
\midrule
\multirow{3}{*}{German} & Train &  4,173,550 &  99,557,517 &  94,741,339 &  881,684 &   1,805,238 &    0.01 &        0.02 \\
      & Dev &      3,000 &      67,807 &      66,412 &    9,778 &      12,859 &    0.14 &        0.19 \\
      & Test &      3,003 &      70,620 &      66,081 &   10,607 &      14,053 &    0.15 &        0.21 \\
\midrule
\multirow{3}{*}{Uzbek} & Train &    529,574 &  11,502,156 &   9,361,833 &  120,768 &     250,629 &    0.01 &        0.03 \\
      & Dev &      2,500 &      52,963 &      42,701 &    8,312 &      13,847 &    0.16 &        0.32 \\
      & Test &      2,500 &      54,061 &      43,945 &    8,349 &      13,265 &    0.15 &        0.30 \\
\bottomrule
\end{tabular}
    }
    \caption{Counts of sentences, tokens and word types in our corpora.}
    \label{tab:corpus_counts_plus_oov}
\end{table*}

Table~\ref{tab:corpus_counts_plus_oov} shows relevant statistics for each translation dataset, including number of sentences, token and type counts, and type-to-token ratios.

\begin{figure*}[t]
    \centering
    \includegraphics[width=\textwidth]{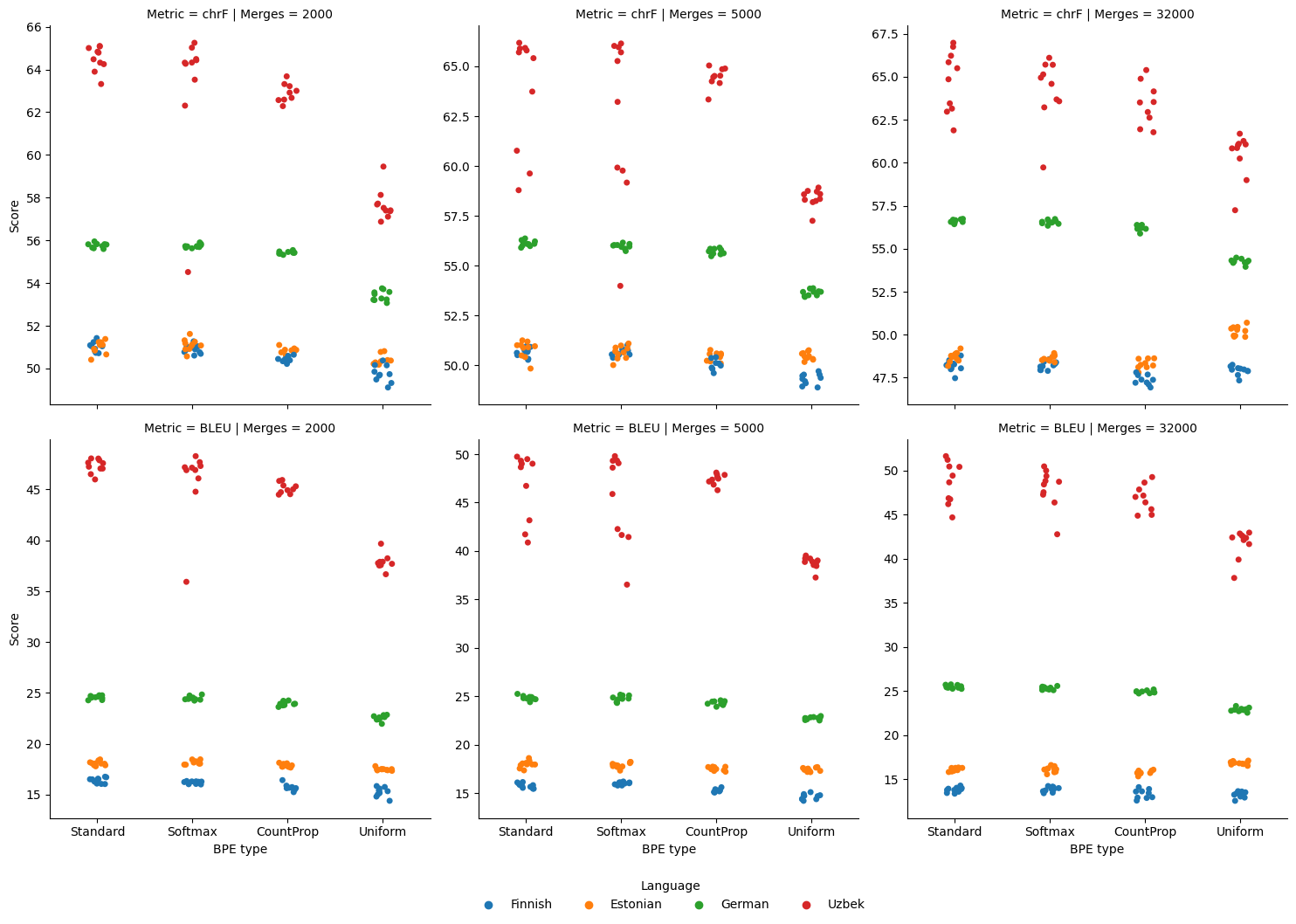}
    \caption{Translation performance across languages and numbers of merges using BLEU (top) and chrF (bottom).}
    \label{fig:main_results}
\end{figure*}

\paragraph{Translation performance} Figure~\ref{fig:main_results} shows a visualization of translation performance in terms of BLEU and chrF across languages and number of merge operations.

\end{document}